# Regressing Relative Fine-Grained Change for Sub-Groups in Unreliable Heterogeneous Data Through Deep Multi-Task Metric Learning

* **Niall O' Mahony, Sean Campbell, Lenka Krpalkova, Joseph Walsh and Daniel Riordan**

Lero – the Irish Software Research Centre, Department of Agricultural and Manufacturing Engineering; School of Science Technology Engineering and Maths (STEM); Munster Technological University – Kerry Campus, Tralee, Co. Kerry, Ireland
E-mail: niall.omahony@research.ittralee.ie

**Abstract:** Fine-Grained Change Detection and Regression Analysis are essential in many applications of Artificial Intelligence. In practice, this task is often challenging owing to the lack of reliable ground truth information and complexity arising from interactions between the many underlying factors affecting a system. Therefore, developing a framework which can represent the relatedness and reliability of multiple sources of information becomes critical. In this paper, we investigate how techniques in multi-task metric learning can be applied for the regression of fine-grained change in real data.

The key idea is that if we incorporate the incremental change in a metric of interest between specific instances of an individual object as one of the tasks in a multi-task metric learning framework, then interpreting that dimension will allow the user to be alerted to fine-grained change invariant to what the overall metric is generalised to be. The techniques investigated are specifically tailored for handling heterogeneous data sources, i.e. the input data for each of the tasks might contain missing values, the scale and resolution of the values is not consistent across tasks and the data contains non-independent and identically distributed (non-IID) instances. We present the results of our initial experimental implementations of this idea and discuss related research in this domain which may offer direction for further research.

## 1. Introduction

Change Detection (CD), the critically important problem of identifying changes in data, constitutes an extensive body of research as there are many applications requiring efficient, effective algorithms for reliably detecting variation. There are many families of CD algorithms that are suitable for different applications [1]. For example, remote sensing applications and the healthcare industry each call for different kinds of algorithms depending on the particular use case. Realtime continuous monitoring by wearables may insist on a simple and efficient statistical algorithm that adapts to the data from each user, while more monitoring more complex systems such as the climate or the environment may call for deep learning and hybrid human-machine intelligence in order to capture all sources of variability [2].

This article focuses on CD with respect to regression problems, where the continuous nature of system scores means they are susceptible to drift and sources of uncertainty due to sensor noise and human error. In these circumstances, the state of the art regression approaches fail to predict subtle deviations for each individual object below the resolution that generalising for the entire population allows. Therefore, instance-based learning approaches are more applicable.

Deep Metric Learning is one such approach where instead of trying to predict a single score/class, the output is mapped to space where you can observe subgroup relatedness. Our has converged towards a Multi-Task Metric Learning (MTML) approach because of the way it learns to map its output to a latent space and how this may be exploited to infer relationships between feature variability and auxiliary background information.

The remainder of this article is organized as follows. In Sect. 2, we briefly introduce notations and algorithms for metric learning. In Sect. 3, we analyse the applicability of metric learning to the problem of fine-grained analysis, in the presence of various sources of uncertainty. In Sect. 4, we review recent advancements in metric learning techniques with respect to each of the challenges presented by our Fine-grained Change Detection application. In Sect. 5, we describe details of some of the techniques we have implemented towards encoding sub-group-specific patterns into latent space with corresponding relative improvements results in regression performance. Lastly, in Sect. 6, we conclude with a brief summary and discussions.

## 2. Deep Metric Learning

In brief, Deep Metric Learning (DML) approaches learn to represent inputs to a lower-dimensional latent space such that the distance between feature vector representations in this space corresponds with a predefined notion of similarity.

DML methods learn a similarity function, where query and gallery data slices are passed pairwise through a Siamese/Triplet/Quadruplet network. I.e. each set is passed through an identical Neural Network, which output features representative of a query input, often called embeddings. The resulting embeddings are compared using some distance metric (e.g. Euclidean distance) and a loss function (such as triplet-loss) is implemented to minimize the distance between embeddings of the same class (inter-class similarity) and maximize the space between classes (intra-class similarity) so that an accurate prediction can be made through querying which is the nearest embedding to a new embedding [3] as depicted in Fig. 1.

One example of a metric loss function is angular loss. Given a triplet contain an anchor, positive and negative $(\mathbf{x}_i, \mathbf{x}_i^+, \mathbf{x}_i^-)$, the angular loss seeks to not only push positive pairs together and negative pairs apart according to distance function $\delta$ as triplet loss does, but also constrains the embeddings to a cosine metric space where the max angle $\angle \mathbf{x}_i^+ \mathbf{x}_i \mathbf{x}_i^-$ denoted as $\alpha$ can be set to make sure negatives are pushed away from the centre of positive clusters (see Fig. 2).

$$[z_i]_+ = \left[\delta_L^2(\mathbf{x}_i, \mathbf{x}_i^+) - 4\tan^2\alpha\, \delta_L^2(\mathbf{x}_i^-, \mathbf{x}_{i-avg})\right]_+ \quad (1)$$

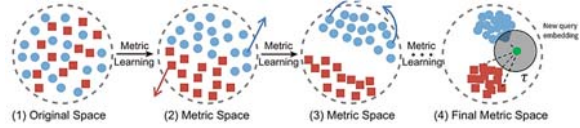

**Fig. 1.** Metric Learning.

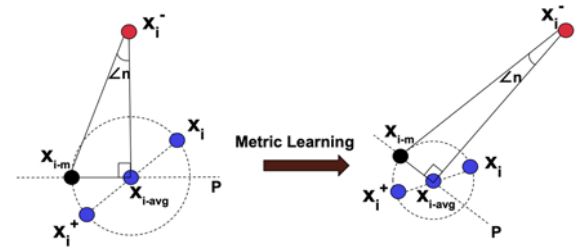

**Fig. 2.** Angular Loss.

The advantages of DML algorithms include: (1) they are very simple and easy to implement; (2) they are usually efficient in space and time complexity; (3) they are often theoretically guaranteed [4]; (4) they are proven to be robust to adversarial attacks [5]. Furthermore, one of the main drives behind recent research in this area is what can be done when we can learn feature embeddings quantifiable in terms of various meaningful metrics, i.e. various transformations, projections and analyses can be performed on the latent space that take advantage human intuition and knowledge from other domains such as calculus of variation and information geometry.

## 3. Applications

The research problems considered by this research consist of the simultaneous learning of multiple tasks and doing so with the additional complexities introduced when analysing real data. Examples of such variable features include seasonal and time-of-day variations in outdoor scenes in place recognition tasks for autonomous navigation and age/gender variations in human/animal subjects in classification tasks for medical/ethological studies [5].

Uncertainty in information retrieval can be introduced by noisy observations and the inability to capture the appearance of all objects from every viewpoint. Sources of noise include; e.g., low resolution/faulty sensors and missing/incorrect human-provided training labels. Examples of uncertainty from uncaptured complexity can be found in many applications (in any system which we do not understand completely in fact). Practical research interests related to modelling uncertainty in complex systems include physics research [6] and robotic perception from limited viewpoints [7].

Adding more labelled data does not eliminate the inherent source of uncertainty. However, modelling uncertainty in training data can improve both the robustness and interpretability of a system [8]. In the remainder of this section, we explore techniques which have been used in conjunction with metric learning to produce accurate predictions even amidst such sources of uncertainty.

### 3.1. Mitigating Uncertainty Due to Sensor Noise

High precision vision/sensing systems for inspection and categorisation of objects are common in modern industrial settings, but only in regulated and optically controlled environments. One of the primary concerns in such systems is accurate change (point) detection (CD), i.e. the identification of the instant at which the dynamics of the underlying system mechanisms change indicating deviation from normal operating conditions. Fine-grained CD (FGCD) is the process of identifying differences in the state of an object or phenomenon where the differences are class-specific and are difficult to generalise. As a result, many recent technologies which leverage big data and deep learning struggle with this task.

### 3.2. Mitigating Uncertainty Due to Human Error

Human-provided reference scores entail subjective assessment which is prone to poor inter-observer and intra-observer reliability. This necessitates for the repeatability of estimation to be compared for \& between observers [6]. Examples of such scenarios include multi-stage disease diagnosis, student satisfaction questionnaire analysis, customer survey analysis [7]. DML methods have also been devised towards handling missing data using a generative model and semi-supervised learning resulting in a less restrictive scenario than the missing completely at random assumption upon which many existing semi-supervised methods are built [8].

### 3.3. Mitigating Uncertainty in Robotics

Robotic perception in unregulated environments is an example of the problem of trying to model complexity beyond which can be conceivably captured in a reference dataset. MTML has been shown to be useful for 3D instance segmentation [9], [10] as it can be used to simultaneously groups parts of the same object instance, i.e. all points in a point cloud segmented as belonging to a particular instance of a class, and estimate the direction towards the instance's centre of mass. This makes separating the object instances much simpler and can also provide a metric for the certainty of the instance proposals as the clusters of vectors should all point to a single point and can also be expected to be of a certain size and distance apart.

Obtaining annotations for scores in regression tasks requires certain degree of domain expertise and hence reference information for these scores is only intermittently available. These annotations are also subject to system errors and subjective biases, which often leads to noisy or erroneous labels. Thus, it is essential to be able to learn a metric that could capture the inherent properties of data, without requiring class labels.

## 4. Related Research

As most CD applications use nominal object classes, many papers applying the latest state-of-the-art to specific application-domains adopt the same methodology, by categorising the output arbitrary classes that ensure even representation across the available dataset. This may come at the cost of not being able to represent the desired output on ordinal, interval or ratio scales which may come naturally in some applications. Thus, they can only measure whether the individual items belong to certain distinct categories and it is not possible to quantify or rank order amongst the categories or perform arithmetic operations. A number of papers have proposed techniques for performing regression with DML [11, 12, 13].

In this section, we review the methods applicable to heterogeneous data from the viewpoint of a number of functionalities, specifically multi-task metric learning, geometry preservation and regression.

### 4.1. Multi-Task Metric Learning

Multi-Task Metric Learning (MTML) approaches jointly train a DML network to separate w.r.t. multiple tasks, thereby sharing useful information among the tasks, which can significantly improve upon the performance achievable in a single-task setting. MTML has been recently reviewed by [14].

Regarding subgroup and fine-grained analyses, Wang, *et al.* propose a regularized multi-task learning framework with shared representation layers to encode the task relatedness [7]. Similarly, [15] present a framework for detecting the subgroups that have similar characteristics in feature space by using K-means clustering incorporated into the regression stage so that both tasks can be performed simultaneously.

It has been proposed in recent work that multiple features should be used for retrieval tasks to overcome the limitation of a single feature and further improve the performance. As most conventional distance metric learning methods fail to integrate the complementary information from multiple features to construct the distance metric, a novel multi-feature distance metric learning method for non-rigid 3D shape retrieval which can make full use of the complementary geometric information from multiple shape features has been presented [16]. An alternative formulation for multi-task learning has been proposed by [17] who use a version of the K Nearest Neighbour (KNN) algorithms (large margin nearest neighbour) but instead of relying on separating hyperplanes, its decision function is based on the nearest neighbour rule which inherently extends to many classes and becomes a natural fit for multi-task learning. This approach is advantageous as the feature space generated from Metric Learning crucially determines the performance of the KNN algorithm, i.e. the learned latent space is preserved, KNN just solves the multi-label problem within.

A challenge that has not been well explored in literature is tuple sampling strategies for metric learning in a multi-label setting. A critical element of metric learning is the selection of triplets during training (sometimes referred to as triplet mining). Triplets need to be chosen such that consecutive training batches vary gradually so that the weights learned by the network between batches are actually transferable but also so that all the important variations within the dataset get encountered with enough frequency during training so that the model can capture that variability. Different mechanisms for triplet mining exist. For example, hard triplet mining selects the most difficult triplets for each anchor by calculating the inter-embedding distance between each anchor and all positive and negative embeddings for that anchor. Hard triplet mining selects the most distant embedding in $S$ and the least distant embedding in $D$ for each anchor. Since it is computationally infeasible to aggregate loss over all $O(n^3)$ triplets and hard triplets can cause models to collapse, heuristics are used to speed up convergence. A well known miner is semi-hard negative mining which samples anchors and positives in batches from $\mathcal{X}$ and $S$, and finds the closest negatives within the batch further away than $D(a,p)$ [18] Many of these sampling strategies can be quite computationally expensive as they require a forward pass of the network being trained to be run on every image in the dataset. This is why many approaches use mini-batches to fit the computational load within the limits of their machine. Many method randomly sample a set of images, assuming that a few random parameter vectors will represent certain elementary transformation operations like translation, scaling, rotation, contrast, and colourization.

### 4.2. Geometry Preserving Metric Learning

The saliency of the embedding space with respect to preserving the local structure of associated tasks/metrics can be increased by taking measures to preserve the intrinsic geometry of the input space. Many of these approaches utilise information geometry divergence measures such as von Neumann divergence [19] and Wasserstein distance [4].

An alternative type of divergence measure is geometric divergence on graphs and manifolds. This can be achieved by fitting manifolds in latent space and including Riemannian metrics in the loss function [12, 20]. Graph-based methods involve representing feature vectors as graphs to capture interactions (i.e. edges) between individual features (i.e. nodes) and are becoming increasingly popular as there has been a massive surge of interest in geometric deep learning in recent years [21].

In their approach called *Orthogonality based Probabilistic Unsupervised Metric Learning* (OPML), Dutta et al. perform Riemannian Conjugate Gradient Descent (RCGD) to jointly learn the parameters [20]. Their implementation of Riemannian optimisation is stochastic, i.e. it samples the latent space at many randomly chosen points and averages the results to obtain a better approximation. This means that the algorithm can be incorporated with deep neural networks for scaling up, and increasing the computational complexity and also the recognition ability of CNNs of work in this area, which up to quite recently was largely purely academic.

The basis of this field of study is that if latent space embeddings are regularised to lie on a manifold then the inferences made embeddings are regularised to have greater consistency and more rigid inter-variable relationships geometrically. This is made possible because manifold structure allows for a number of operations to be carried out which are advantageous in fine-grained change regression applications. The first is that operations on manifolds can be exploited for regularising the distribution of embeddings such that the model can be taught to have the divergence of embedding clusters be indicative of fine-grained shift in score labels. A Riemannian manifold, or simply a manifold, can be described as a continuous set of points that appears locally Euclidean at every location. This property i often referred to as smoothness. Every smooth manifold has a Riemannian metric. A Riemannian metric (tensor) makes it possible to define several geometric notions on a Riemannian manifold, such as angle at an intersection, length of a curve, area of a surface and higher-dimensional analogues (volume, etc.), extrinsic curvature of sub-manifolds, and intrinsic curvature of the manifold itself.

While the most familiar manifold is Euclidean space, a large number of other, more exotic manifolds can be defined. The manifolds discussed in this paper are examples of matrix manifolds, meaning that they consist of points that can be represented as matrices. More specifically, a manifold is a topological space and a collection of differentiable, one-to-one mappings called charts. At any point on a $d$ dimensional manifold, there is a chart that maps a neighbourhood containing that point to the Euclidean space $R^d$. This property allows a tangent space to be defined at every point, which is an Euclidean space consisting of directions that point along the manifold. Tangent spaces are particularly important for optimization because they characterize the set of update directions that can be followed without leaving the manifold [22].

## 5. Implementation

We implement our experiments on a custom image dataset related to animal health monitoring of 228 animals containing 9 annotations for each image. Some of the annotation labels (3) are up to date at the time of image capture as they are fed from an automated system, while some (2) are provided by human experts on an intermittent basis (every 3 weeks) and missing values were replaced with the closest observation. The remaining 4 values indicate the forward and backward gradient of theses expert-provided scores.

The problems to be solved share common challenges around the recognition of subgroup features in dynamic unregulated environments for which include the need to recognise objects never seen before during training, with non-IID instances, in real-time, on limited hardware, in such a way to align dependent information and predict various different types of data.

In working through these problems we have realised a unified metric learning approach which incorporates multi-task and geometry preserving metric learning and have experimented with applying the following components to a deep metric/manifold learning framework [20] trained to perform an multi-label identification task and then being made to infer an auxiliary score and instance-instance change for said score.

The motivation for adopting the methodology of [20] is that enforcing orthogonality between multiple tasks allows natural clusters in the data to be detected, while also simultaneously learning from manual/system annotations where provided. Our application requires that the *intra-class variances* that may occur in our dataset, e..g. due to shift in animal health indication scores whether it be gradual and natural over a period or sudden. Such variances can be captured by adding an orthogonality constraint to the Mahalanobis matrix used for distance estimation. This constraint ensures these relationships are apparent while also allowing the push-pull nature of triplet/angular metric loss function to distribute embeddings that fall between class centres appropriately relying on notions of geometric similarity within the data.

In contrast to Dutta's application, where $\mathcal{X}$ is labelled, we guide our triplet mining strategy on multiple labels. Regarding subgroup and fine-grained analyses, we propose a regularized multi-task learning framework with shared representation layers to encode the task relatedness. A class label matrix was generated to be stored with the image data. The training dataset is randomly sampled for anchors and then iteratively sample for positives and negatives as follows:

Let $\mathbf{x}_i \in \mathbb{R}^d$ be the descriptor of an example $i$ in a dataset $\mathcal{X}$, which is labelled with class label matrix $\mathbf{y}_i$. To form a set of triplets: $\mathcal{T} = \{(\mathbf{x}_i, \mathbf{x}_i^+, \mathbf{x}_i^-)\}_{i=1}^{|\mathcal{T}|}$, each element of which consists of the following:

1) $\mathbf{x}_i$, an arbitrary example with $(\mathbf{y}_i)$, which denotes the class label matrix assigned to $\mathbf{x}_i$, $\mathbf{x}_i^+$, another arbitrary example with $(\mathbf{y}_i^+) = (\mathbf{y}_i)$ for at minimum $n$ elements.

2) $\mathbf{x}_i^-$, such that $(\mathbf{y}_i^-) = (\mathbf{y}_i)$ for less than $n$ elements. The examples $\mathbf{x}_i$, $\mathbf{x}_i^+$ and $\mathbf{x}_i^-$ are referred to as the *anchor*, *positive* and *negative* respectively.

| Jumbo | Lactation | Calving Date | BCS@calving | 18/01/2018 | 02/02/2018 | 16/02/2018 | 07/03/2018 | 22/03/2018 | 06/04/2018 | 19/04/2018 | 04/05/2018 | 18/05/2018 | 31/05/2018 | 19/06/2018 |
|---|---|---|---|---|---|---|---|---|---|---|---|---|---|---|
| 170 | 3 | 20/02/2018 | 3.25 | | | | | | | | | | | |
| 172 | 3 | 25/02/2018 | 3 | | | | | | | | | | | |
| 325 | 3 | 05/02/2017 | 4.75 | 4.5 | | | | | | | | | | |
| 360 | 3 | 11/09/2017 | 3.5 | 3.5 | 3.5 | 3.5 | | | | | | | | |
| 523 | 5 | 08/09/2017 | 3.25 | 4 | | | | | | | | | | |
| 570 | 6 | 01/02/2018 | 3.25 | 3.25 | 3 | 2.75 | 2.5 | 2.5 | 2.75 | 2.75 | 2.75 | 2.75 | 2.75 | 2.5 |
| 603 | 4 | 15/03/2017 | | | | | | | | | | | | |
| 610 | 4 | 20/09/2017 | | | | | | | | | | | | |
| 790 | 5 | 06/03/2018 | 3.25 | 3 | 3 | 3 | 3.25 | 3.25 | 3 | 3 | 3 | 3.25 | 3.25 | 3.25 |
| 796 | 7 | 24/03/2017 | | | | | | | | | | | | |

**Fig. 3.** Our Multi-Task Triplet Mining Strategy: If n of the animal health attribute labels (circled in diagram, details not important) match up for any pair, then they are a positive pair. Note that some values may be missing in the dataset.

For $\mathbf{x}_i \in \mathbb{R}^d$, let $\mathbf{L}^\top \mathbf{x}_i \in \mathbb{R}^l$, denote its learned embedding. Our goal is to learn $\mathbf{L}$ in $\delta_\mathbf{L}^2(\mathbf{x}_i, \mathbf{x}_j)$. Let, the learned embeddings of the anchor and the positive be denoted as $\mathbf{L}^\top \mathbf{x}_i$ and $\mathbf{L}^\top \mathbf{x}_i^+$ respectively. The bilinear similarity between them can be expressed as $(\mathbf{L}^\top \mathbf{x}_i)^\top \mathbf{L}^\top \mathbf{x}_i^+ = \mathbf{x}_i^\top \mathbf{L} \mathbf{L}^\top \mathbf{x}_i^+$.

$$\min_{\mathbf{R},\mathbf{L}} \mathcal{L}(\mathbf{R}, \mathbf{L}) = \sum_{i=1}^{|\mathcal{T}|} (-\log(p_i))$$

Here, $\mathbf{L} \in \mathbb{R}^{d \times l}$ is the parametric matrix of the squared Mahalanobis-like distance metric $\delta_\mathbf{L}^2(\mathbf{x}_i, \mathbf{x}_j) = (\mathbf{x}_i - \mathbf{x}_j)^\top \mathbf{L} \mathbf{L}^\top (\mathbf{x}_i - \mathbf{x}_j)$, for a pair of examples $\mathbf{x}_i, \mathbf{x}_j \in \mathbb{R}^d$. Ensuring $l < d$ facilitates dimensionality reduction. **R** is a parametric metrix that facilitates a weight funcion based on similarity. $\sum_{i=1}^{|\mathcal{T}|}(-\log(p_i))$ is the log likelihood of the triplet set $\mathcal{T}$ satisfying the angular constraint. The goal is to learn the parametric matrix **L**.

Our novel multi-task triplet mining strategy has a number advantages when combined with manifold optimisation techniques. Firstly, because negative pairs can arise from pairs with one or many mismatching variablres, if tuned appropriately, the training process may learn to enforce orthogonality of some of the latent dimensions in correspondence with the difference in the ground truth labels. Secondly, because there is a degree of randomness to the level of mismatch of the input variables, somewhat similar pairs are going to be matched more often than those with few variables matching, hence that similarity is mapped to the latent manifold at a broader scale.

Similarly to [7], we also implement shared representation layers to encode the task relatedness but differ in that we simply calculate the mean squared error (MSE) of a selected dimension of the feature embedding with the label of interest (generalised regression and relative change scores respectively).

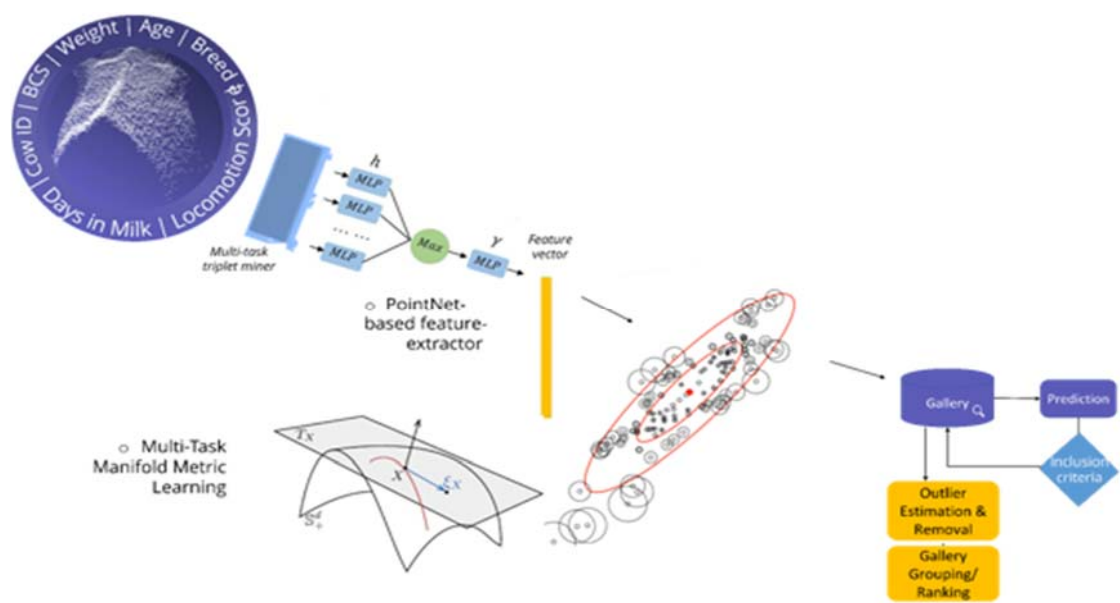

**Fig. 4.** An overview of our approach, the experiments in this article focus on the middle portion where malanobis distance (shown in red) has a orthogonality preserving constraint [20] which, in coordination with our Multi-Task Triplet Mining Strategy enforces that instances with some deviant variables be orthogonal and hence the deviation is made quantifiable in latent space.

## 6. Results

We perform experiments to study the following:
1) Effect of different sampling settings on the multi-task metric learning results.
2) Effect of the different hyperparameters on our method.

In Table 1, we report the results of the multi-task triplet sampler. For our method, we set α=45°, and embedding size of 128. We also indicate the nature of an approach, i.e., we use the OPML network.

The test of sensitivity to n, the number of elements in the class label matrix which must match to qualify as a positive sample yielded interesting results. When n is small, the labels which match can vary significantly within and between batches and hence the transition between training iterations is difficult to learn from. This gradually improves as n increases as is reflected in Table 1. Once greater than 6 of the 9 labels in the data needed to match, performance dropped again, suspect due to the limited number of anchor-positive pairs accepted.

**Table 1.** Sensitivity of the multi-task triplet sampler and OPML network towards n, the number of elements in the class label matrix which must match to qualify as a positive sample.

| n Network | 2 | 3 | 4 | 5 | 6 | 7 |
|---|---|---|---|---|---|---|
| **OPML** | 71.2 | 62.3 | 50.2 | 42.5 | 47.0 | 61.6 |

In Table 2, we report the results of α, an important hyperparameter in OPML and the addition of a MSE component to the overall loss function. For our method, we arbitrarily set n=5, and embedding size of 128. We also indicate the nature of an approach, i.e., whether it makes use of an additional final layer for classification and regression or not. Note: lower loss is better.

55

**Table 2.** Sensitivity towards loss function and α
with respect to loss values reached on the test data.

| loss \ α | 35° | 40° | 45° | 50° | 55° | 60° |
|---|---|---|---|---|---|---|
| Multi-task/ OPML | 73.4 | 70.5 | 69.5 | 68.7 | 71.7 | 72.4 |
| Multi-task/ OPML+MSE | 241.2 | 162.3 | 142.2 | 152.5 | 92.0 | 321.6 |

These results corresponds with ablation studies by [20] showing the effect of different components of OPML in that a value of α in the mid-range of that tested gave the best results, where α is the upper-bound angle formed by the angle positive-anchor-negative in the angular loss function, however the loss values remained quite high suggesting that the model is not converging well. This may be due to the complexity of the base network compared to that used by [20].

## 7. Conclusions

The multiple implementation scenarios where MTML solutions have been shown to be useful demonstrate that the proposed method is useful in many AI applications where the algorithm's outputs are to be adapted to each use case.

This paper has also included the results of implementing some of these methods in addition to a novel multi-task triplet sampling strategy to quantify the effect of the hyperparameters of each element. It was found that enforcing too many labels to overlap impeded the convergence of the loss function as there was a limited number of anchor-positive pairs in the dataset. Allowing very little overlap of labels was also detrimental to performance as it allows too much variance between training iterations.

Adding a final layer to regress the embeddings to align to a set of auxiliary outputs hindered the training procedure which may have been due to over-constraining the loss function, to have to learn to space the vectors apart appropriately while simultaneously enforcing a feature which may have no or non-linear correlation to the auxiliary task. To this end, some of the kernel regression techniques [23] may improve performance.

Better results may be achieved if a metric loss function be used in the final layer and the intra-class distances are trained to correspond to the auxiliary tasks using the weighted MSE loss function. The benefit of the interaction of multiple features is to be included in efforts to minimise the regression error as it is why methods which use a Mahalanobis matrix for each metric are so common in MTML solutions.

We intend to further the research in this area by investigating how the mapping element of MTML may be exploited in situations where the salient features vary over time or due to changing underlying variables. One possible avenue would be to develop a technique to calculate the geometric divergence along each of the dimensions of the latent space/surface w.r.t isolated task-relevant subgroups in order to identify the local covariate geometry with respect to each task and easily recognise subtle changes in the same.

## Acknowledgements

This work was supported, in part, by Science Foundation Ireland grant 13/RC/2094 and co-funded under the European Regional Development Fund through the Southern \& Eastern Regional Operational Programme to Lero - the Irish Software Research Centre (www.lero.ie). The authors wish to acknowledge the DJEI/DES/SFI/HEA Irish Centre for High-End Computing (ICHEC) for the provision of computational facilities and support.

## References

[1]. W. Shi, M. Zhang, R. Zhang, S. Chen, Z. Zhan, Change detection based on artificial intelligence: State-of-the-art and challenges, *Remote Sensing*, Vol. 12, Issue 10, May 2020, p. 1688.

[2]. N. O' Mahony, S. Campbell, L. Krpalkova, A. Carvalho, J. Walsh, D. Riordan, Representation learning for fine-grained change detection, *Sensors*, Vol. 21, Issue 13, Jul. 2021.

[3]. N. O' Mahony, *et al.*, One-shot learning for custom identification tasks; a review, in *Procedia Manufacturing*, Vol. 38, 2019, pp. 186-193.

[4]. C. Shui, M. Abbasi, L. É. Robitaille, B. Wang, C. Gagné, A principled approach for learning task similarity in multitask learning, in *Proceedings of the IJCAI International Joint Conference on Artificial Intelligence*, Vol. 2019, 2019, pp. 3446–3452.

[5]. L. Wang, X. Liu, J. Yi, Y. Jiang, C. J. Hsieh, Provably robust metric learning, *Advances in Neural Information Processing Systems*, Vol. 33, 2020, pp. 1-12.

[6]. Z. Wang, D. M. Anand, J. Moyne, D. M. Tilbury, Improved sensor fault detection, isolation, and mitigation using multiple observers approach, *Systems Science and Control Engineering*, Vol. 5, Issue 1, 2017, pp. 70–96.

[7]. L. Wang, D. Zhu, Tackling ordinal regression problem for heterogeneous data: sparse and deep multi-task learning approaches, *Data Mining and Knowledge Discovery*, Vol. 35, Issue 3, 2021, pp. 1134-1161.

[8]. X. Liu, D. Zachariah, J. Wågberg, T. B. Schön, Reliable Semi-Supervised Learning when Labels are Missing at Random, *arXiv*, Issue 2, 2018, pp. 1–8.

[9]. W. Wang, R. Yu, Q. Huang, U. Neumann, SGPN: Similarity Group Proposal Network for 3D Point Cloud Instance Segmentation, in *Proceedings of the IEEE Computer Society Conference on Computer Vision and Pattern Recognition*, 2018, pp. 2569-2578.

[10]. J. Lahoud, B. Ghanem, M. R. Oswald, M. Pollefeys, 3D instance segmentation via multi-task metric learning, in *Proceedings of the IEEE International Conference on Computer Vision*, Vol. 2019-Octob, 2019, pp. 9255-9265.

[11]. K. Q. Weinberger, G. Tesauro, Metric learning for kernel regression, *Journal of Machine Learning Research*, Vol. 2, Issue May, 2007, pp. 612-619.

[12]. R. Huang, S. Sun, Kernel regression with sparse metric learning, *Journal of Intelligent and Fuzzy Systems*, Vol. 24, Issue 4, 2013, pp. 775-787.
[13]. A. Taha, Y.-T. Chen, T. Misu, A. Shrivastava, L. Davis, Unsupervised Data Uncertainty Learning in Visual Retrieval Systems, *arXiv*, 2019.
[14]. P. Yang, K. Huang, A. Hussain, A review on multi-task metric learning, *Big Data Analytics*, Vol. 3, Issue 1, Dec. 2018, pp. 1–23.
[15]. B. Liang, P. Wu, X. Tong, Y. Qiu, Regression and subgroup detection for heterogeneous samples, *Computational Statistics*, Vol. 35, Issue 4, 2020, pp. 1853-1878.
[16]. H. Wang, H. Li, J. Peng, Xianping Fu, Multi-feature distance metric learning for non-rigid 3D shape retrieval, *Multimedia Tools and Applications*, Vol. 78, Issue 21, Nov. 2019, pp. 30943-30958.
[17]. K. Q. Weinberger, L. K. Saul, Distance metric learning for large margin nearest neighbor classification, *Journal of Machine Learning Research*, Vol. 10, Jan. 2009, pp. 207-244.
[18]. R. Manmatha, C. Y. Wu, A. J. Smola, P. Krahenbuhl, Sampling Matters in Deep Embedding Learning, in *Proceedings of the IEEE International Conference on Computer Vision*, Vol. 2017-Octob, 2017, pp. 2859-2867.
[19]. P. Yang, K. Huang, C. L. Liu, Geometry preserving multi-task metric learning, *Machine Learning*, Vol. 92, Issue 1, Jul. 2013, pp. 133-175.
[20]. U. K. Dutta, M. Harandi, C. C. Sekhar, Unsupervised Deep Metric Learning via Orthogonality Based Probabilistic Loss, *IEEE Transactions on Artificial Intelligence*, Vol. 1, Issue 1, 2020, pp. 74-84.
[21]. M. M. Bronstein, J. Bruna, Y. LeCun, A. Szlam, P. Vandergheynst, Geometric Deep Learning: Going beyond Euclidean data, *IEEE Signal Processing Magazine*, Vol. 34, Issue 4, 2017, pp. 18-42.
[22]. S. Giguere, F. Garcia, S. Mahadevan, A Manifold Approach to Learning Mutually Orthogonal Subspaces, *arXiv*, 2017.
[23]. K. Q. Weinberger, G. Tesauro, Metric Learning for Kernel Regression, in *Proceedings of the 11th International Conference on Artificial Intelligence and Statistics (AISTATS'2007)*, San Juan, Puerto Rico, March 21-24, 2007, pp. 612-619.

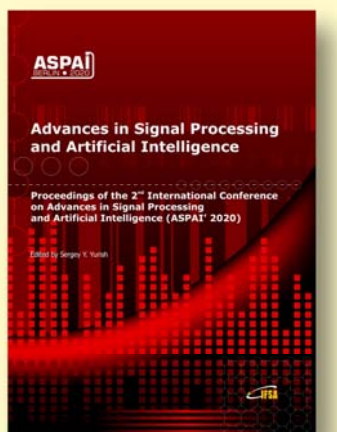
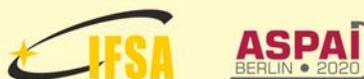
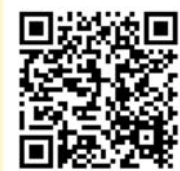